\documentclass[conference]{IEEEtran}
\IEEEoverridecommandlockouts
\usepackage{cite}
\usepackage{amsmath,amssymb,amsfonts}
\usepackage{algorithmic}
\usepackage{graphicx}
\usepackage{textcomp}
\usepackage{xcolor}
\usepackage{siunitx}
\usepackage[utf8]{inputenc}
\def\BibTeX{{\rm B\kern-.05em{\sc i\kern-.025em b}\kern-.08em
T\kern-.1667em\lower.7ex\hbox{E}\kern-.125emX}}

\begin{document}
\raggedbottom
\title{Recognizing Emotions evoked by Movies using Multitask Learning\\
\thanks{978-1-6654-0019-0/21/\$31.00 ©2021 European Union}}

\author{\IEEEauthorblockN{Hassan Hayat, Carles Ventura, Agata Lapedriza}
\IEEEauthorblockA{\textit{Estudis d'Informatica, Multimedia i Telecomunicacio / eHealth Center} \\
\textit{Universitat Oberta de Catalunya, Barcelona, Spain}\\
\{hhassan0,cventuraroy,alapedriza\}@uoc.edu}
}

\maketitle

\begin{abstract}
Understanding the emotional impact of movies has become important for affective movie analysis, ranking, and indexing. Methods for recognizing evoked emotions are usually trained on human annotated data. Concretely, viewers watch video clips and have to manually annotate the emotions they experienced while watching the videos. Then, the common practice is to aggregate the different annotations, by computing average scores or majority voting, and train and test models on these aggregated annotations. With this procedure a single aggregated evoked emotion annotation is obtained per each video. However, emotions experienced while watching a video are subjective: different individuals might experience different emotions. In this paper, we model the emotions evoked by videos in a different manner: instead of modeling the aggregated value we jointly model the emotions experienced by each viewer and the aggregated value using a multi-task learning approach.
Concretely, we propose two deep learning architectures: a Single-Task (ST) architecture and a Multi-Task (MT) architecture. Our results show that the MT approach can more accurately model each viewer and the aggregated annotation when compared to methods that are directly trained on the aggregated annotations. Furthermore, our approach outperforms the current state-of-the-art results on the COGNIMUSE benchmark.
\end{abstract}


\section{Introduction}

\IEEEPARstart{U}{nderstanding} the emotional experience of users when watching a movie has been broadly studied in psychology \cite{russell2003core,plantinga2013affective,plantinga2008emotion} and has recently attracted the attention among computer scientists \cite{2016mediaeval,sjoberg2015mediaeval,soleymani2014corpus,thao2019multimodal,Poria_2017}. Models that can automatically infer the emotions evoked by movies are useful in multiple applications and scenarios, such as recommendation systems, affective multimedia retrieval, or for producers to understand the impact of the content they create. 

One of the challenges to create automatic systems to infer emotions evoked by movies is the subjective nature of the task: the same content can evoke different emotions to different people \cite{muszynski2019recognizing}. This challenge has not been explicitly studied before in the context of emotions evoked by movies, and it is actually the focus of our work. In general, works on evoked emotion recognition focus on modelling aggregated annotations of multiple viewers, such as the average \cite{Thao_2019,timar2018feature,liu2019multimodal,pini2017modeling}. However, our results show that, given a crowd of viewers, jointly modelling the perception of each viewer and the average across viewers in a multi-task manner can actually produce more accurate results than just modelling the average viewer in a single-task manner. The intuition behind this idea is the following. Each viewer has different sensitivities and preferences, which affect the emotion experienced while watching a movie. Thus, to model the evoked emotion for a specific viewer, a machine learning system needs to find the patterns in the input data (the movie) that are informative for inferring the evoked emotion for that viewer. However, when we aggregate the experienced emotion of different viewers, the viewer specific patterns get merged, making the patterns of the input data to infer the aggregated annotation harder to find. On the contrary, with the multi-task approach that we suggest, the model needs to find the patterns corresponding to each viewer to model each viewer. Thus, the representation of these viewer specific patterns will be encoded in the latent representation of the CNN, and will be available for the classification branch of the aggregated annotations, making the modelling of the aggregated annotations easier.

\begin{figure}[htb]
  \begin{center}
  \includegraphics[width=\linewidth]{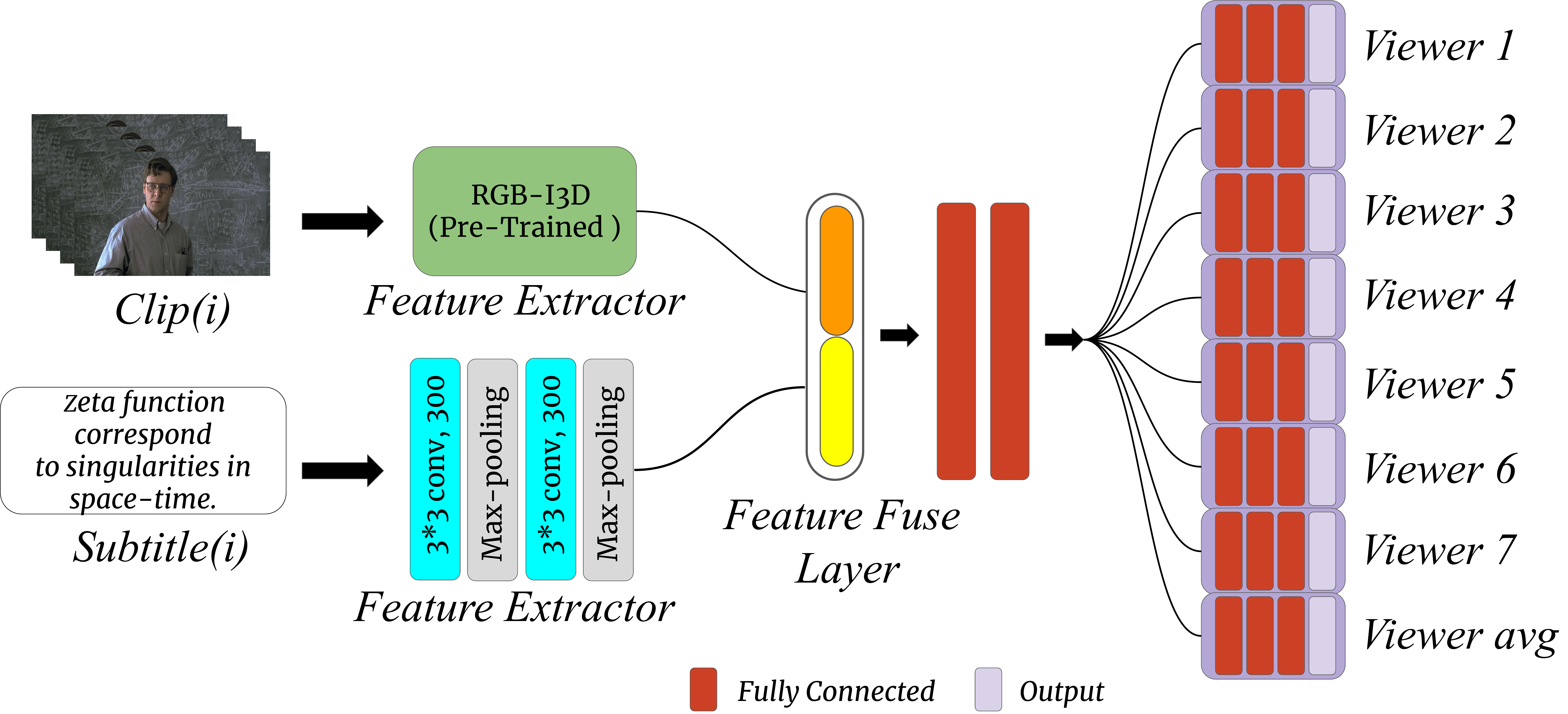}
  \caption{Architecture of the Multi-Task model. This architecture is applied to all viewers in a joint manner by sharing a common two fully-connected layers and a dedicated fully-connected block for each viewer. The weights of the shared convolutional block, and two fully-connected layers are updated in every training batch whereas the weights of the fully-connected block are only updated when the batch contains annotations of that specific viewer. }
  \label{fig:mutli_task}
  \end{center}
\end{figure}

In this paper we propose a Multi-Task Deep Learning architecture (MT) to jointly model the perception of each viewer and the aggregated annotation of the average viewer, as illustrated in Fig.\ref{fig:mutli_task}. Concretely, our  MT architecture is an extension of our Single-Task formulation (ST), since it uses the same backbone modules for the feature extraction of the text and visual modalities as in the ST architecture. However, after the fusion layer we add two fully connected layers, and then we create separate classification branches, one per each of the viewers and one more for the aggregated annotation. The details of our architectures are described in Sect.\ref{sec:emotion_modelling}. 


To validate our approach we conduct experiments on the COGNIMUSE dataset \cite{zlatintsi2017cognimuse}. In particular we use the standard benchmark for evoked emotion understanding, which includes clips of $7$ Hollywood movies. The COGNIMUSE benchmark is annotated on evoked valence, arousal and dominance by $7$ different viewers. In this work we focus on the valence dimension, which is the most popular dimension for emotion recognition related tasks \cite{zheng2017identifying,becker2017emotion,wang2019sentidiff,rosas2013multimodal,chang2017learning}. For our study we follow the common practice of using two discrete values \cite{Nguyen2019,ghaleb2019multimodal,ko2018brief}, i.e. positive vs. negative. As discussed in Sect.\ref{related work}, COGNIMUSE is the only publicly available dataset for evoked emotion recognition in movies that provides the annotations of separate viewers, instead of just an aggregated value, which is the key requirement for our study. More details about the dataset and the data partitions used in our experiments are provided in Sect.\ref{results}.

Our experiments are presented in Sect.\ref{sec:results_main}. First, we compare our proposed ST and MT models. For that we create a collection of data splits, as described in Sect. \ref{sec:data-folds}, and we report the average accuracy obtained by ST and MT models on each viewer as well as on the average viewer. In this case we observe that the MT obtains higher accuracies than the ST. Then, we perform ablation experiments on the separate modalities, and observe again that the MT outperforms the ST for both the text and the visual modalities. Furthermore, our results of the ablation study are consistent with previous findings showing higher accuracies for the visual modality when compared to the text modality. Finally, we compare our approach with the results reported in \cite{Nguyen2019}, which are the current state-of-the-art results for valence classification in the COGNIMUSE benchmark. The modelling approach proposed in \cite{Nguyen2019} is referred as Baseline in our paper and it is briefly described in Sect.\ref{sec:baseline_model}. As shown in Table \ref{tab:results_ablation_baselinesplit}, we observe that our proposed MT approach outperforms the Baseline model.

In summary, the results of our work suggest that, for the problem of evoked emotion recognition, jointly modelling each viewer and the average viewer can be a better solution than just modelling the average viewer in a single task manner. 
\section{Related Work}
\label{related work}
Recognizing emotions evoked by movies has been getting more attention due to its potential applications in affective based recommender or affective multimedia retrieval systems \cite{Poria_2017,jan2016bul,ma2016thu}, among others. For example, Bao et al \cite{bao2013your} used smartphone sensors to detect the viewer's emotional cues, such as laughter, during the movie, while Lee et al. \cite{lee2014emotion} used 3d fuzzy visual and EEG features to understand the emotional state of viewers while watching a movie. More recently, Thao et al. \cite{thao2019multimodal} used audiovisual data for predicting affective responses of viewers watching a movie, while Nguyen et al.\cite{Nguyen2019} proposed a multimodal approach that uses audiovisual and text data to understand the emotions evoked by movies. The authors in \cite{Nguyen2019} proposed different Convolutional Neural Netwokrs (CNNs) per each of the modalities considered in their work (visual, text, and audio). In this work the architectures of \cite{Nguyen2019} are used as a baseline to compare our results. Sect.\ref{sec:baseline_model} gives a deeper explanation on this baseline approach.

One of the challenges of evoked emotion recognition is that evoked emotion is subjective, meaning that the same content can evoke different emotions to different viewers. Our approach in this paper is to jointly learn the individual experienced emotion and the aggregated experienced emotion in a Multi-task manner to better approximate both the individual and the aggregated experiences. However, the problem of dealing with subjective annotations has been recently addressed by \cite{ye_subjective_annotations} in a different way. The authors proposed a probabilistic approach to estimate reliability and regularity of viewers in order to aggregate annotations. Reliability computes how likely the viewer's response is serious (instead of random), while regularity measures how the viewer agrees with the other viewers. Then, these reliability and regularity scores are used to compute a final aggregate value per each instance, which is considered the ground truth. More generally, subjectiveness-aware methods are getting more and more attention due to the fact that the established model outperforms human raters on average in terms of the perception uncertainty \cite{liu2019multimodal,pini2017modeling}. 
On the other hand, \cite{fayek2016modeling} proposed user-level emotion modeling of different viewers and then combined individual annotations using geometric mean and unweighted majority voting for final predictions. The performance gain using user-level modeling is better than using the ground truth labels obtained by averaging or majority voting directly. Similarly, \cite{chou2019every} used the majority of annotations (hard label) and the distribution of annotations (soft label) simultaneously for joint emotion modeling. Later they concatenated all the outputs and used Softmax for the final predictions. Also, \cite{li2019attentive} proposed a personalized emotion recognition model that integrated different user’s attributes for emotion classification. To address subjectivity \cite{han2017hard} introduced user-dependent weight to predict the final emotional state.  

A key component to train automatic systems for evoked emotion recognition is data. In this work we use the COGNIMUSE dataset \cite{zlatintsi2017cognimuse}. COGNIMUSE is the only dataset for evoked emotion recognition in movies that provides the annotations of each individual viewer (the data used in our study is explained in more depth in Sect. \ref{dataset}). However, there are other labeled datasets for evoked emotion in videos. For example, the HUMAINE dataset \cite{douglas2007humaine} consists of $50$ video clips, having $0.5$ - $3$ minutes in length. Each clip was annotated at two levels: Global level (emotions related state, context labels, key events, etc) and frame-by-frame (intensity, arousal, valence, dominance). FilmStim \cite{schaefer2010assessing} has $70$ movies with the length ranging from $1$ to $7$ minutes. It uses unique global labels for emotion ranking (subjective arousal, positive and negative affect, six emotions discreteness score, etc). LIRIS-ACCEDE database \cite{baveye2015liris} contains $9,800$ videos excerpts, having a range from $8$ to $12$ seconds, and annotated in $2D$ (arousal, valence) domain. Unfortunately these datasets just provide an aggregate annotation per instance, such as the average or the majority voting, which makes them not suitable for our study.

\subsection{Baseline Model}
\label{sec:baseline_model}
The CNN-based architectures proposed in \cite{Nguyen2019} are considered as the Baseline models in our experiments. In particular, we compare our results with the multi-modal approach from \cite{Nguyen2019}, as well as with the separate text and visual modalities, referred as Baseline-Text and Baseline-Visual, respectively. 

The Baseline-Text consists of three convolution layers and each layer has $100$ kernels, which extract the high-level text features for classification. Each convolution layer is followed by a ReLu activation function, a max-pooling layer, and a Fuzzified kernel. The words from the sentence were pre-processed before passing through the layers of the convolutional neural network. A maximum length of $18$ words was considered and a particular word was padded at the end of shorter sentences.  

The Baseline-Visual has three convolutional layers having $64$, $96$, and $128$ kernels, respectively. The input of the CNN is the fusion of two different features: the orientation features, which were calculated based on \ang{0}, \ang{45}, \ang{90}, \ang{135}; and the color features, which were obtained by converting RGB to HSI color space. The final features are constructed by computing a $32$-bin histogram of each $7$ channels ($4$ from orientation and $3$ from color). 

\section{Proposed Emotion Modeling}
\label{sec:emotion_modelling}
In our experiments we use two types of CNN architectures: a Single-Task (ST) architecture and a Multi-Task (MT) architecture. Both architectures use the same backbone modules for the text and the visual modalities, respectively. 

\subsection{Single-Task architecture (ST)}
\begin{figure}[htb]
  \begin{center}
  \includegraphics[width=91mm]{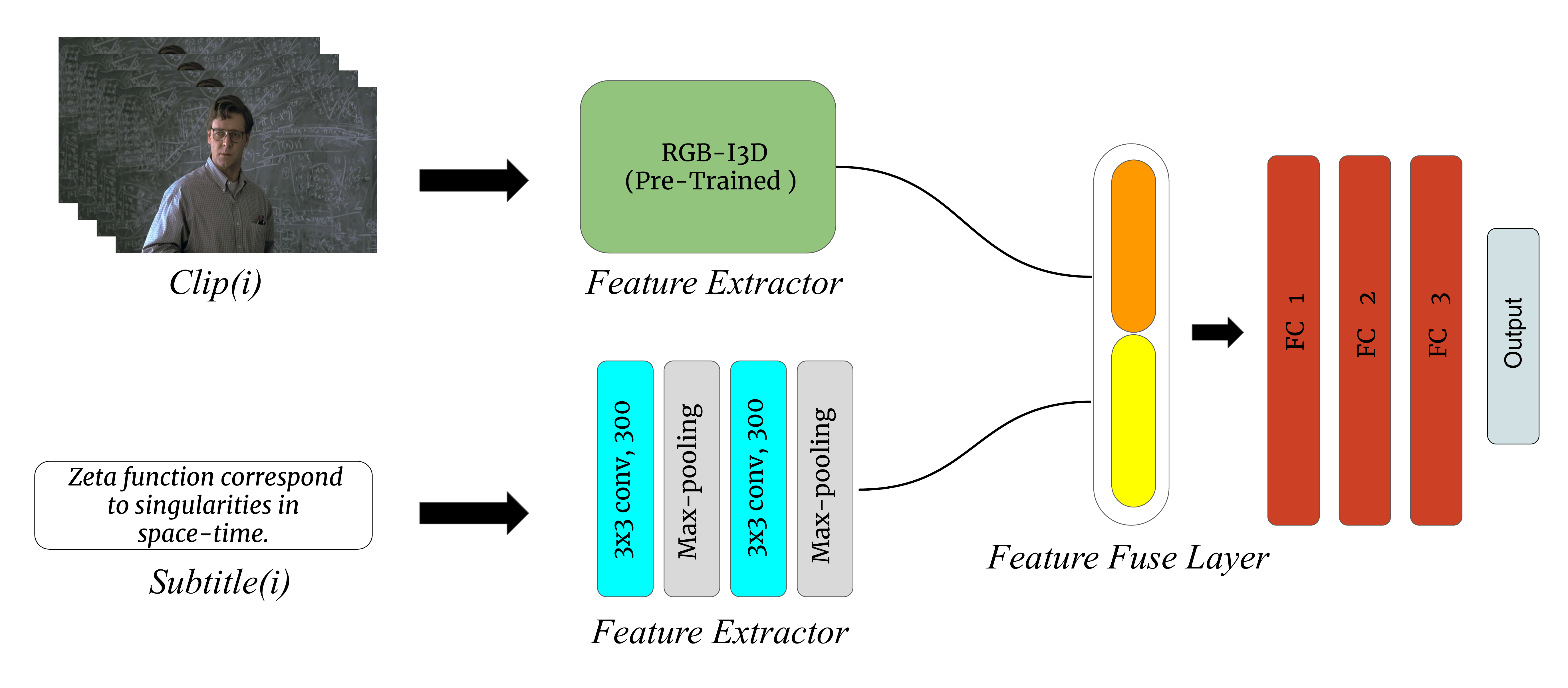}
  \caption{Architecture of the Single-Task (ST) model. It consists of feature extractors, feature fusion layer, and fully-connected block.}
  \label{fig:single_task}
  \end{center}
\end{figure}

Our ST architecture is illustrated in Fig.\ref{fig:single_task}. The architecture has two input branches, one per each modality (visual -top branch- and text -bottom branch-). Per each branch we use the corresponding backbone to extract the features (the backbone modules are described below in this section). Then, the features are fused by concatenation. After that, our architecture has three fully connected layers with $1024$, $512$, and $256$ units, respectively. 
The sigmoid function is applied to the output layer to get the final prediction. Due to the high difference between the numbers of positive and negative labels in each viewer's dataset (see Fig.\ref{Data distribution}), we used the weighted log-loss to compensate for the imbalanced data. This way, the contribution to the loss of both positive and negative examples is the same regardless of the distribution of the labels in the training dataset. Adam optimizer \cite{kingma2014adam} with learning rate $10^{-3}$ is used for training. Finally, to overcome overfitting we use a  $Lasso$ regularization term \cite{Tibshirani1996}.

\begin{figure}[htb]
  \begin{center}
  \includegraphics[width=90mm]{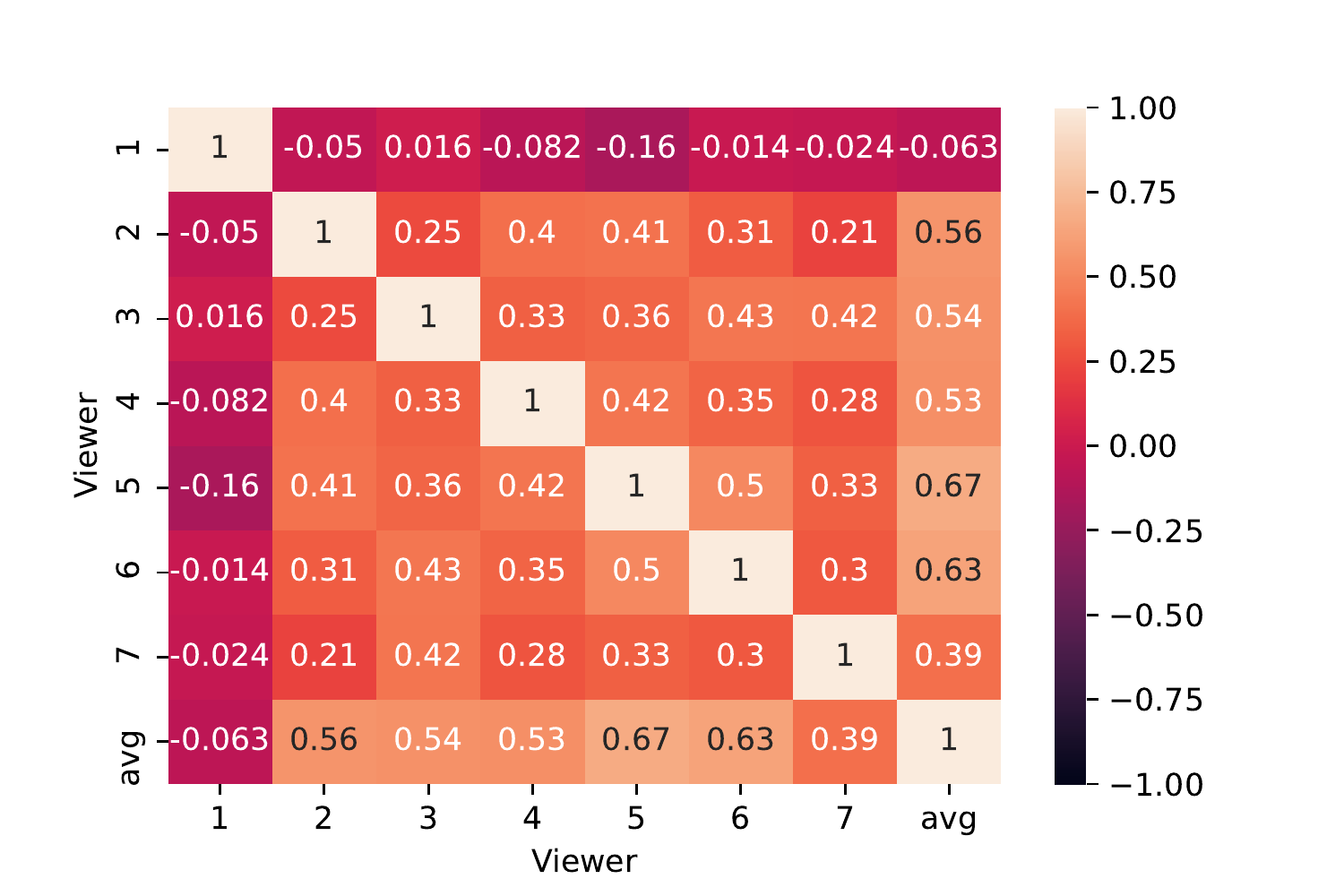}
  \caption{Average correlation matrix of all movies. Correlation values are computed for each pair of viewers as well as for each viewer with respect to the aggregated (averaged) annotations. It can be observed that the correlation values between different viewers is low. However, the correlation value between each viewer and the aggregated annotations is significantly higher.}
  \label{Average correlation}
  \end{center}
\end{figure}

\subsection{Multi-Task architecture (MT)} 
\label{Multitask}

Our MT architecture is illustrated in Fig.\ref{fig:mutli_task}. In this case we use the same backbone feature extraction branches for the visual and the text modality as in the ST architecture. Then we concatenate the features and add two fully connected layers of $2048$ and $1024$ units, respectively. After that we have separate branches per each viewer, including all $7$ viewers and the average viewer. Each of the separate branches have 3 fully connected layers of $512$, $256$, and $128$ units, respectively. Finally we use a sigmoid and log loss, again with a $Lasso$ regularizer. 
Notice that, as illustrated in Fig.\ref{fig:mutli_task}, we have one separate output branch per each one of the 7 viewers ($V_{1}$- $V_{7}$), as well as a last branch for the average viewer ($V_{avg}$). 

\subsection{Text backbone branch} 
\label{Backbone_text}

Our backbones for the text and visual modalities are illustrated in both Fig.\ref{fig:single_task} and Fig.\ref{fig:mutli_task}. For the text modality we learn a word-embedding matrix to map every word in a sentence to a d-dimensional vector. This way, sentences can be represented as vectors of numerical values. Since the length of the sentences is variable, a maximum length of 18 words is considered and a particular word is padded at the end of shorter sentences. As a result, all the sentences have the same length. After that we use a sequence of two pairs of convolutional layer plus max pooling. We initialize this text feature extraction branch randomly and we train it with the labeled data. 

\subsection{Visual backbone branch} 
\label{Backbone_visual}

For the visual modality, we used a fixed pre-trained RGB-I3D model \cite{Carreira_2017} on the Kinectic-400 Dataset. The architecture uses $3$D convolutions and max-pooling operations to learn seamless spatio-temporal features from video. The I3D model is known as the Inflated 3D that is based on the state-of-the-art Inception-V1 \cite{ioffe2015batch} model. All the convolutional and pooling filters of Inception-V1 were converted from $2$D into $3$D. This additional dimension is known as the temporal dimension and helps the model in learning temporal patterns of the video. Each convolutional layer is followed by batch normalization \cite{ioffe2015batch} and a ReLU activation function. The I3D model has multiple end-points to collect the features of the given input video. In our experiments, we used the features of the last endpoint of the model that is “Mixed-5c". We processed a batch of $16$ consecutive frames with a stride of $8$ frames of a single clip and the features are then global average-pooled. To get the most important segments of the clip we then max-pooled across the temporal domain. 

\section{Data} \label{results}
\label{dataset}

In our experiments we use the COGNIMUSE dataset \cite{zlatintsi2017cognimuse}. COGNIMUSE is a collection of movie clips and travel documentary clips with human annotations on different tasks: audio-visual and semantic saliency, audio-visual events and action detection, cross-media relations, and emotion recognition. In our study we use data from the emotion recognition benchmark. This section describes the data used in our study and the data split generated for our experimental protocol. We also provide at the end of this section a brief statistical analysis of the data. 

\subsection{COGNIMUSE: Evoked emotions}
\label{dataset description}
The standard benchmark for emotional understanding \cite{zlatintsi2017cognimuse} includes $7$ Hollywood movies: \emph{“A beautiful Mind” \textbf{(BMI)}, “Chicago” \textbf{(CHI)}, “Crash” \textbf{(CRA)}, “Finding Nemo” \textbf{(FNE)}, “Gladiator” \textbf{(GLA)}, “The Departed” \textbf{(DEP)}, and “Lord of the Rings - the Return of the King” \textbf{(LOR)}} and each movie has $30$ minutes in length. The emotions evoked by the movies are represented in the valence and arousal space. Valence represents how positive or negative the emotion evoked by the clip is, while Arousal encodes viewer’s excitement, agitation, or readiness to act. The frame rate of each movie is 25 fps and 7 different viewers provided annotations for each frame in continuous values from $-1$ to $+1$ of arousal and valence domains. In this work we focus on the Valence dimension.

The dataset also includes the movie subtitles of each video segment (notice that the total number of video frames is different for each video segment). We consider each video segment as a sample for our model, where the input is the text associated to that video segment and the output is the averaged valence value along the time dimension. This output value results from averaging all valence values annotated (we have a value every $40$ ms). Once the averaged valence value is obtained, it is binarized using the value 0 as a threshold. The values greater than the threshold are considered as positive, whereas the values below the threshold are considered as negative. Using this discretization for valence recognition is a common practice, as we can also see in \cite{Nguyen2019}.

\subsection{Data distribution analysis}
\label{sec:viewer_correlations}

In this section we analyze the distribution of the data and the agreement among the different viewers. First, Fig.~\ref{Average correlation} shows the correlations between the different viewer pairs. We can observe that viewer 1 has very low correlations with the other viewers. Furthermore, we notice that several correlation values below $0.4$. We also observe that some viewers are more correlated to the average viewer than others. For example, viewer 5 is highly correlated with the average viewer, viewer 7 is poorly correlated with the average viewer, and viewer 1 has almost 0 correlation with the average viewer.

\begin{figure*}[htb]
\centering
\begin{tabular}{@{}cccc@{}}
\includegraphics[width=.23\textwidth]{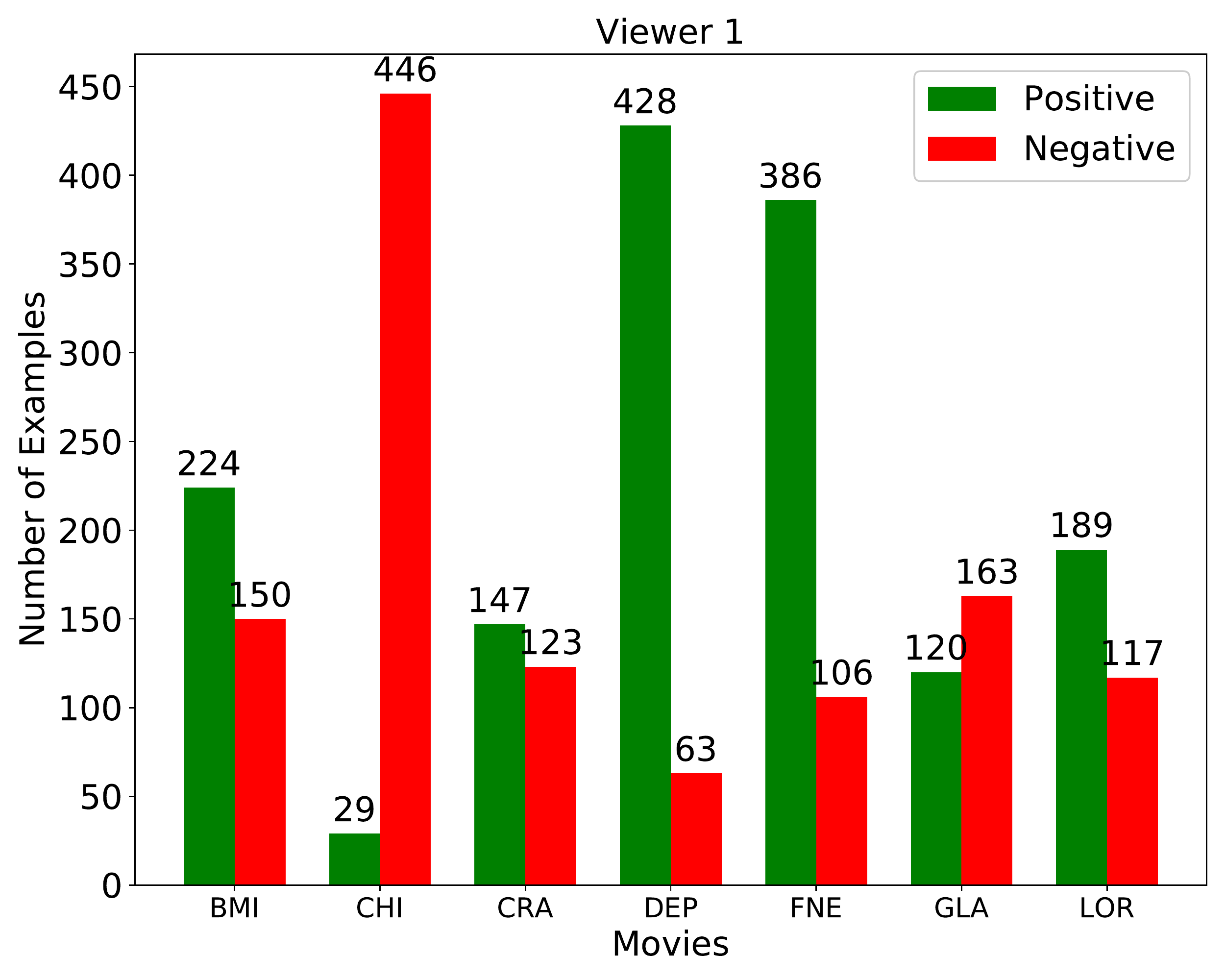} &
\includegraphics[width=.23\textwidth]{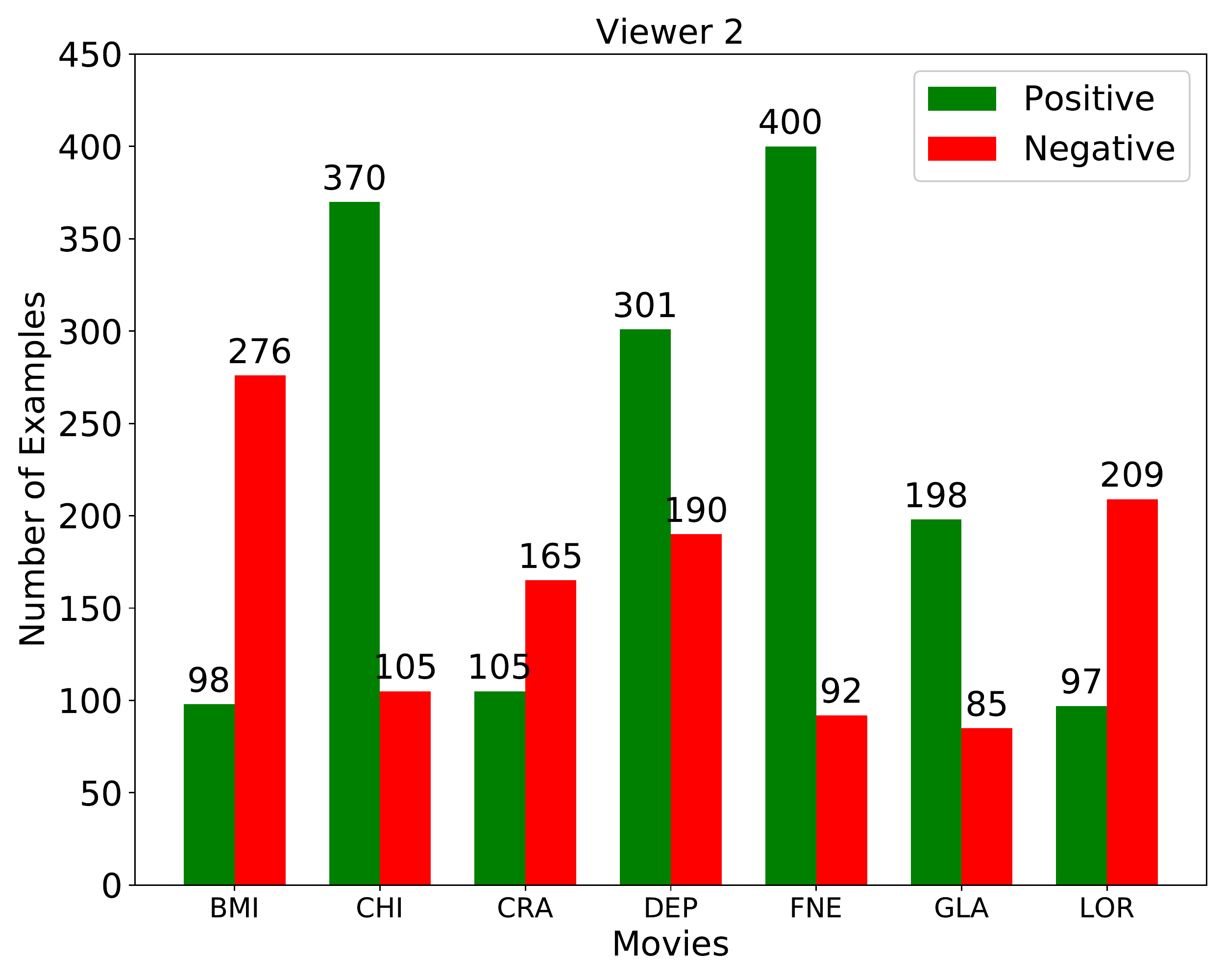} &
\includegraphics[width=.23\textwidth]{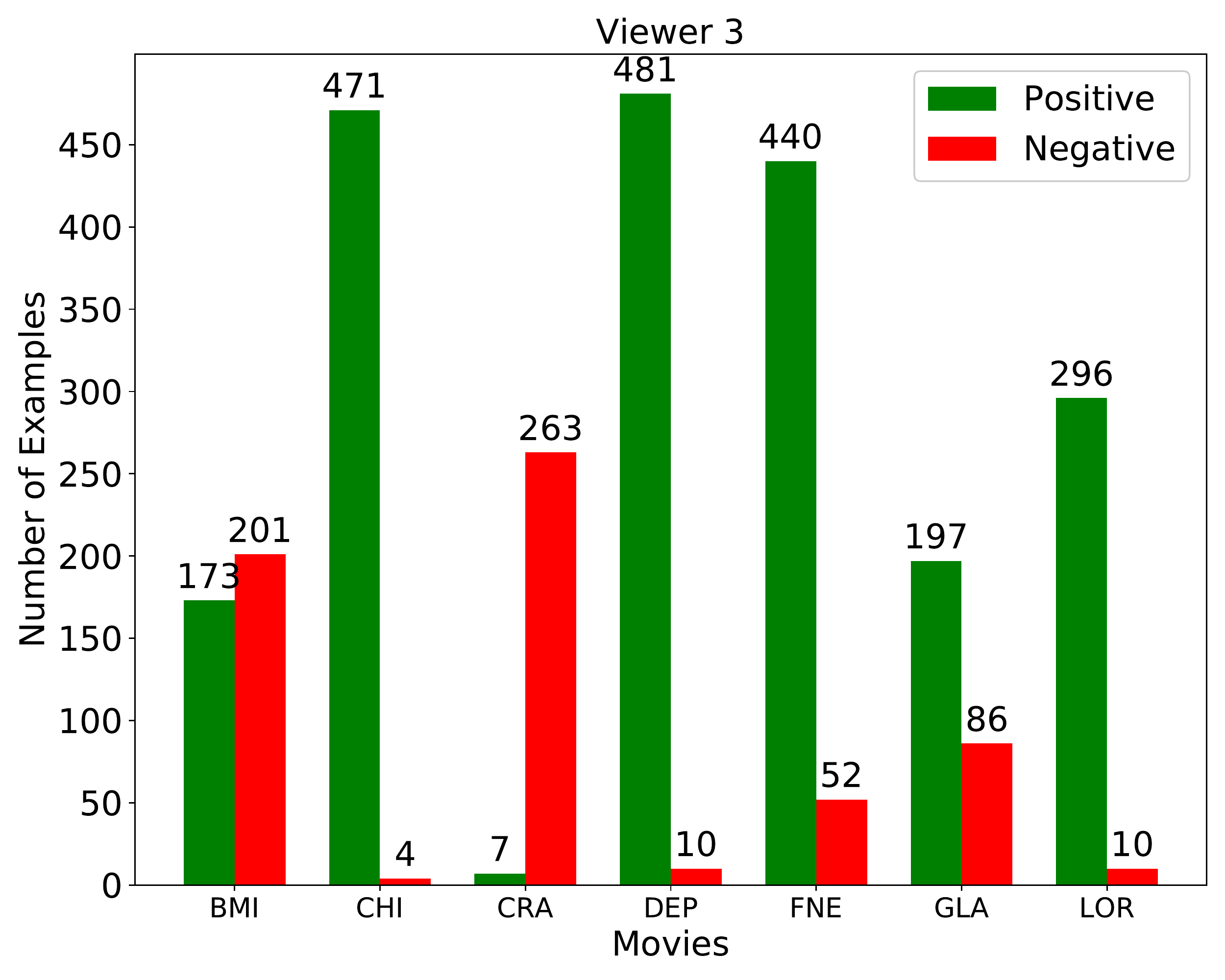} &
\includegraphics[width=.23\textwidth]{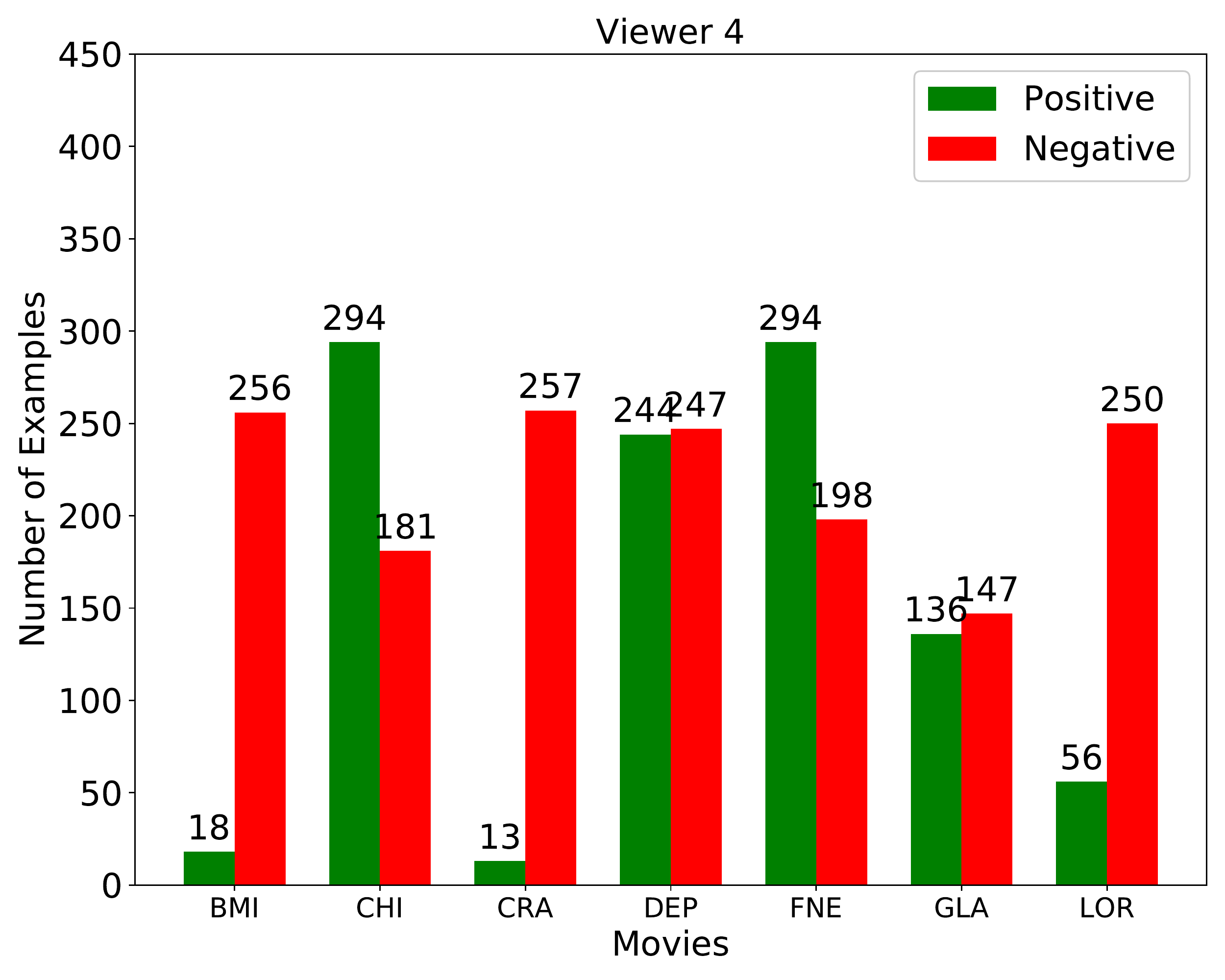} \\
\includegraphics[width=.23\textwidth]{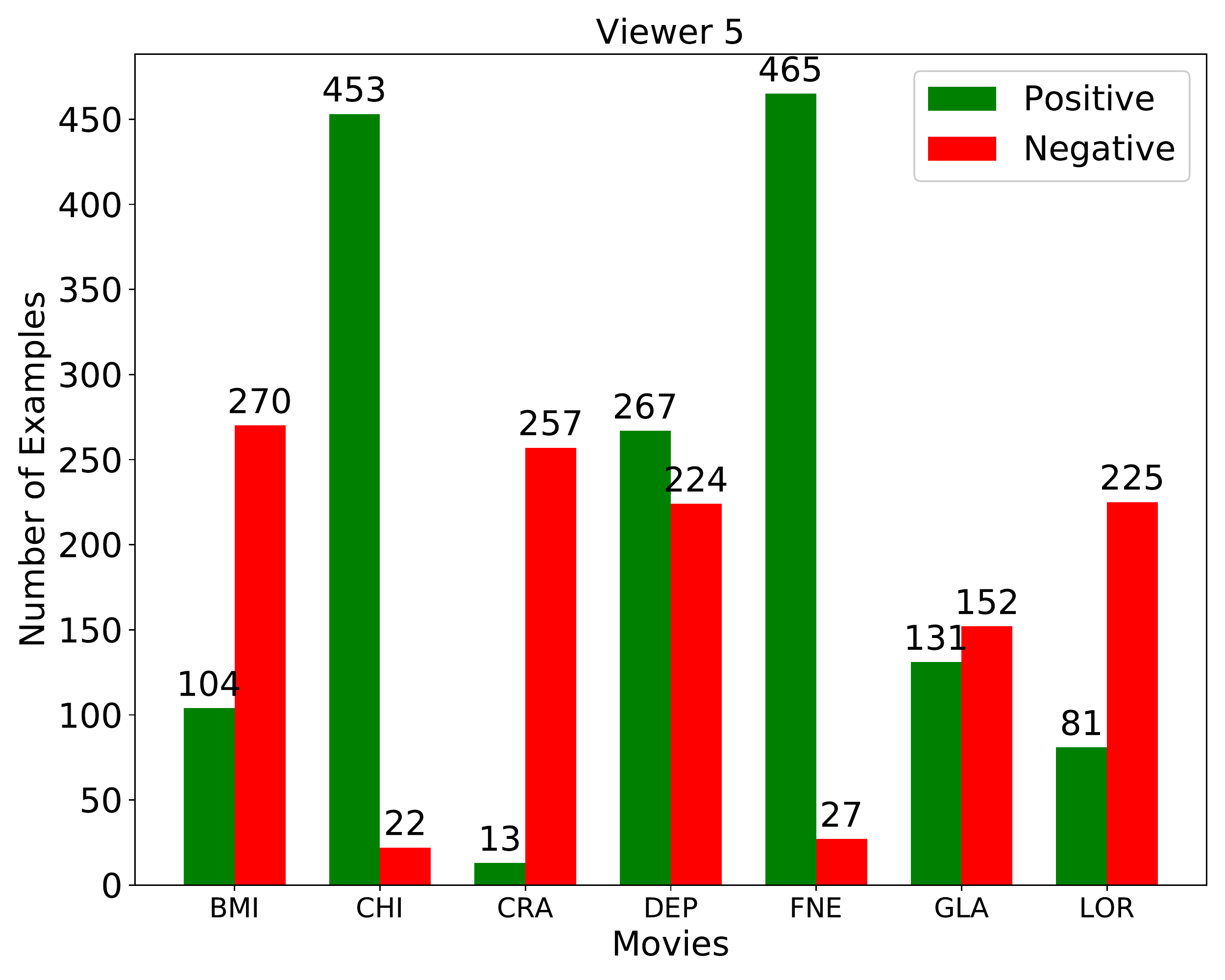} &
\includegraphics[width=.23\textwidth]{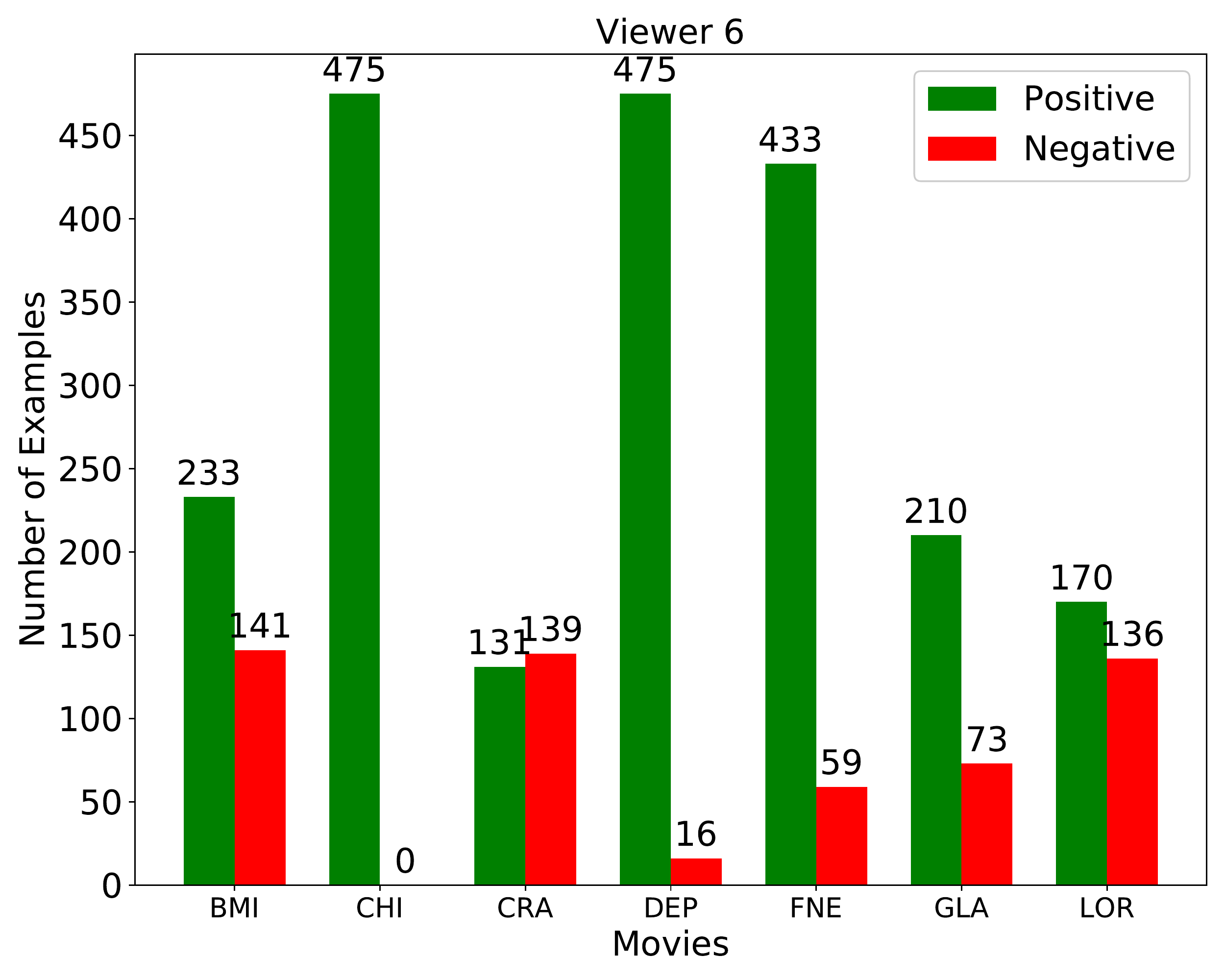} &
\includegraphics[width=.23\textwidth]{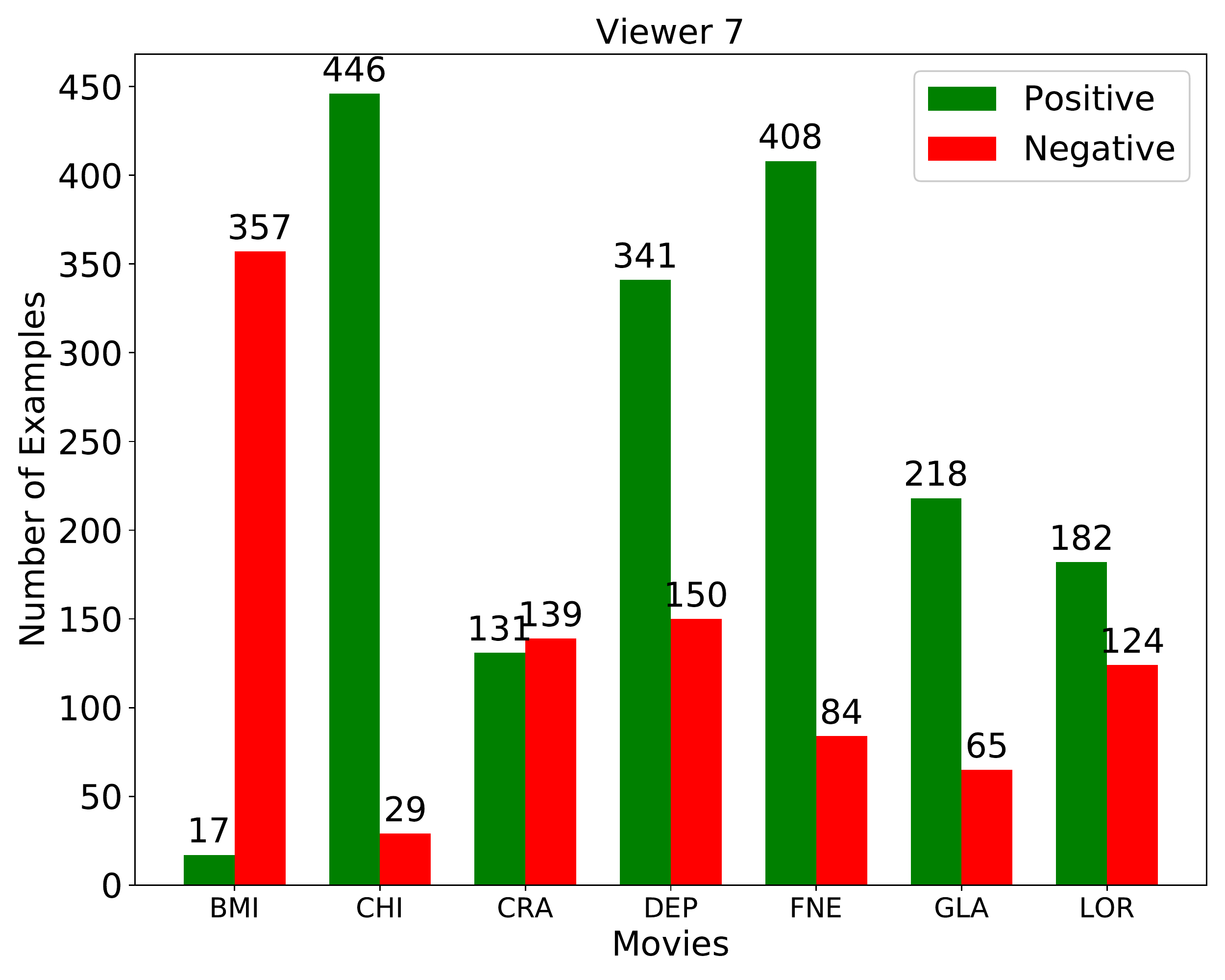} &
\includegraphics[width=.23\textwidth]{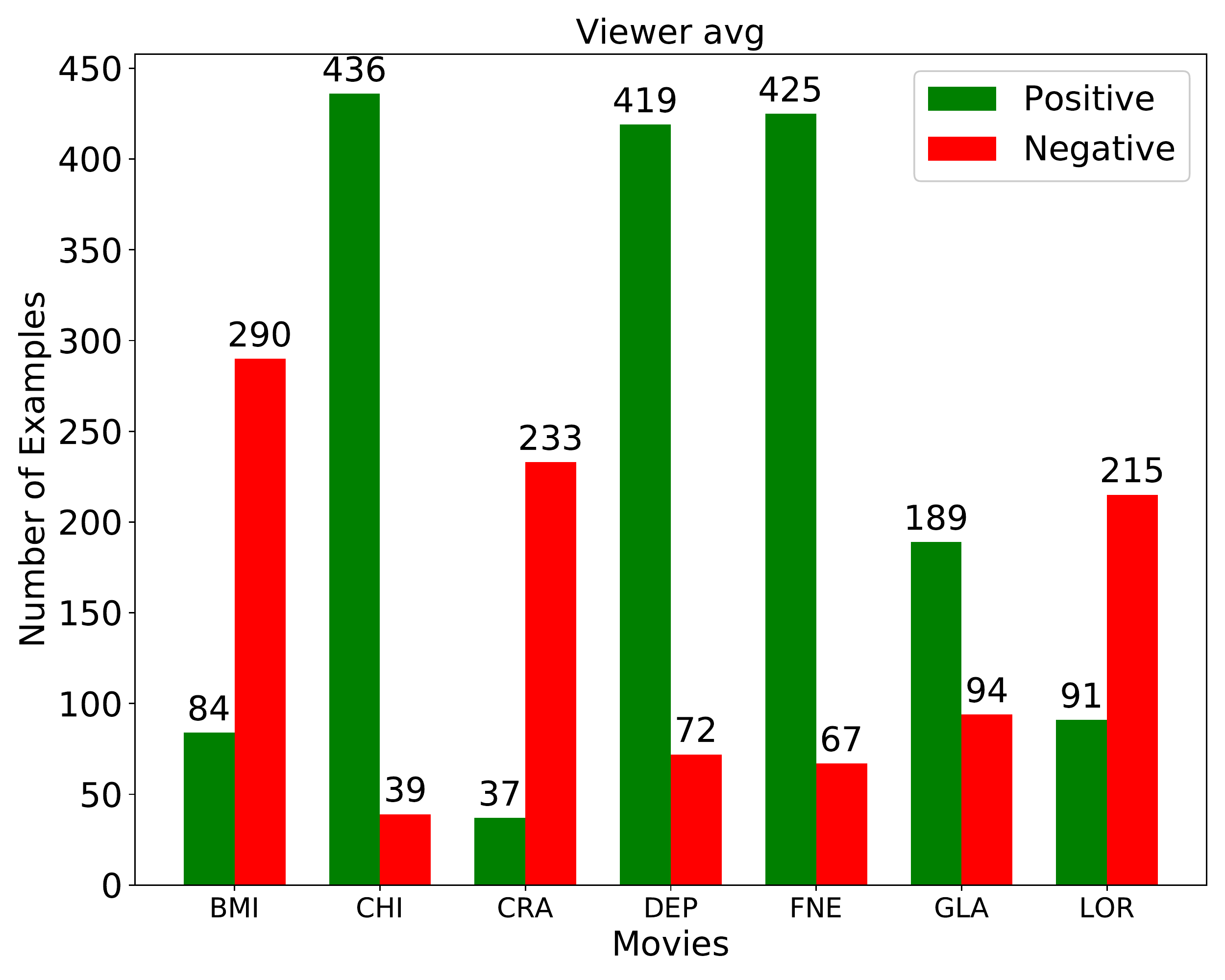} \\
\end{tabular}
\caption{Viewer annotation distribution for Valence (Positive vs. Negative) on each movie, including the average viewer (last plot). Notice that, for all the movies, the distribution of positive and
negative labels significantly varies across the different viewers.}
\label{Data distribution}
\end{figure*}

Fig. \ref{Data distribution} shows the histogram (total counts) of positive and negative labels per viewer and per movie. Again, we observe notable differences among the viewers. In particular, we can observe how the histogram of viewer 1 is very different from the other histograms, and how the histogram of viewer 5 and average viewer are the most similar ones. 

Overall, these statistics illustrate the subjectiveness of the viewers: they often experience different emotions when viewing the same movie segment. 

\subsection{Data folds}
\label{sec:data-folds}


We define seven different folds for cross-validation to perform an accurate evaluation of the proposed models. 
For each fold, one movie is left for test and the remaining six movies are randomly split into training (5 movies) and validation (1 movie). For reproducibility purposes, Table~\ref{cross-validation folds} summarizes the 7 data folds used in our study, which are denoted by $F_1,...,F_7$. 

\begin{table}[hbt!]
 \caption{Detail of the data folds. Movies used for training, validation and testing are specified for each fold.}
  \begin{center}
    \begin{tabular}{|c|c|c|c|} 
      \hline
      \textbf{Fold id} & \textbf{Training} & \textbf{Validation} & \textbf{Testing}\\ [0.5ex] 
      \hline\hline
      $F_1$ & CRA, DEP, FNE, GLA, LOR & CHI & BMI \\
      \hline
      $F_2$ & BMI, DEP, FNE, GLA, LOR & CRA & CHI \\
      \hline
      $F_3$ & BMI, CHI, FNE, GLA, LOR & DEP & CRA \\
      \hline
      $F_4$ & BMI, CHI, CRA, GLA, LOR & FNE & DEP \\
      \hline
      $F_5$ & BMI, CHI, CRA, DEP, LOR & GLA & FNE \\
      \hline
      $F_6$ & BMI, CHI, CRA, DEP, FNE & LOR & GLA \\
      \hline
      $F_7$ & BMI, CRA, DEP, FNE, GLA & CHI & LOR \\
      \hline
     \end{tabular}
    
     \label{cross-validation folds}
  \end{center}
\end{table}

\section{Experiments and Results} 
\label{results}

We perform extended experiments to compare the different approaches for modelling the emotion evoked by movies. First, Sect.\ref{sec:results_main} shows the results obtained when using the ST and MT multi-modal approaches to model each viewer, including the average viewer. Second, Sect.\ref{sec:results_ablation} presents our ablation study, where we show the results obtained by each separate modality (i.e. text and visual). Then, we compare our models with the Baseline model \cite{Nguyen2019} and, finally we discuss the obtained accuracies and show qualitative results.  

\subsection{ST and MT Multi-Modal architectures}
\label{sec:results_main}

\begin{table*}[hbt!]
\caption{Accuracies obtained when modelling each viewer using the Multi-Modal architectures. As a reference, we provide accuracies obtained by a random classifier, a Positive classifier (i.e. a classifier that always classifies any instance as positive), and a Negative classifier (i.e. a classifier that always classifies any instance as negative). $V_{avg}$ denotes the average viewer, while Mean (last column) is the mean accuracy, i.e. average of all the columns.}
    \begin{center}
    \begin{tabular}{|c|c|c|c|c|c|c|c|c|c|} 
      \hline
       & $V_1$ & $V_2$ & $V_3$ & $V_4$ & $V_5$ & $V_6$ & $V_7$ & $V_{avg}$ & Mean \\ 
      \hline\hline
      Random & 50.00 & 49.90 & 50.00 & 50.00 & 49.90 & 
      50.00 & 49.90 & 50.00 & 49.62\\
      \hline
      Positive & 57.00 & 58.00 & 77.00 & 39.00 & 56.00 & 79.00 & 65.00 &  62.00 & 61.62 \\
      \hline
      Negative & 43.00 & 42.00 & 23.00 & 61.00 & 44.00&
      21.00 & 35.00 & 38.00 & 38.37  \\
      \hline \hline
      ST (ours) & 63.56 & 63.73 & 62.80 & 69.81 &
      65.50 & 69.46 & 64.60 & 65.72 & 65.64 \\
      \hline
      MT (ours) & \textbf{67.60} & \textbf{71.40} & \textbf{69.50} & \textbf{73.20} & \textbf{82.60}&
      \textbf{75.35}& \textbf{73.64} & \textbf{77.80}&
      \textbf{73.61}\\
      \hline
      \end{tabular}
      
      \label{res:multi_modal_viewer}
      \end{center}
\end{table*}

\begin{table*}[hbt!]
\caption{Results of the ablation studies. Accuracies obtained when modelling each viewer using just the Text modality.}
    \begin{center}
    \begin{tabular}{|c|c|c|c|c|c|c|c|c|c|} 
      \hline
       & $V_1$ & $V_2$ & $V_3$ & $V_4$ & $V_5$ & $V_6$ & $V_7$ & $V_{avg}$ & Mean \\
       \hline\hline
       Baseline-Text \cite{Nguyen2019} & 45.30 & 50.77 & \textbf{64.80} & 48.50 &
       45.20 & 67.50 & 42.45 & 48.40 & 51.74 \\
      \hline\hline
      ST-Text (ours) & 54.60 & 51.30 & 51.72 & 52.66 & 53.40 &
      59.88 & 48.50 & 51.70 & 52.96 \\
      \hline
      MT-Text (ours) & \textbf{56.86} & \textbf{64.50} & 
      64.73 & \textbf{69.55} & \textbf{60.41}&
      \textbf{67.88} & \textbf{66.26} & \textbf{70.16}&
      \textbf{65.04}\\
      \hline
      \end{tabular}
      \label{tab:results_ablation_text}
      \end{center}
\end{table*}

\begin{table*}[hbt!]
\caption{Results of the ablation studies. Accuracies obtained when modelling each viewer using just the Visual modality.}
    \begin{center}
    \begin{tabular}{|c|c|c|c|c|c|c|c|c|c|} 
      \hline
       & $V_1$ & $V_2$ & $V_3$ & $V_4$ & $V_5$ & $V_6$ & $V_7$ & $V_{avg}$ & Mean \\
     \hline\hline
      Baseline-Visual \cite{Nguyen2019} & 53.56 & 50.61 & 61.45 & 55.40 &
      68.74 & 57.80 & 57.60 & 54.52 & 57.46  \\
      \hline\hline
      ST-Visual (ours) & 61.50 & 61.73 & 61.66 & 66.30 &
      62.85 & 66.40 & 62.35 & 64.70 & 66.60 \\
      \hline
      MT-Visual (ours) & \textbf{67.25} & \textbf{69.70} &
      \textbf{68.50} & \textbf{67.55} & \textbf{80.40}&
      \textbf{65.77} & \textbf{66.90} & \textbf{74.80}&
      \textbf{70.10}\\
      \hline
      \end{tabular}
      \label{tab:results_ablation_visual}
      \end{center}
\end{table*}

Table \ref{res:multi_modal_viewer} shows the results obtained when modelling each viewer with the ST model vs. the MT model. Concretely, each column corresponds to one viewer, $V_i$, and shows the average accuracies obtained when approximating $V_i$ for the different data folds, $F_j$. We denote by $V_{avg}$ the average viewer, computed by averaging the annotations of all the $7$ individual viewers. As a reference, the first three rows of the table show the results obtained by a random classifier, a Positive classifier (i.e. assigning a positive output to any input instance), and a Negative classifier (i.e. assigning a negative output to any input instance). To facilitate the comparisons we add in the last column (denoted by Mean) the average result per row. 

As expected, we observe that the MT models obtain higher accuracies in average, both in the case of the individual viewers (65.64 vs. 73.61) as well as for the average viewer (65.72 vs. 77.80). These results are coherent with other studies in the context of Affective Computing, that also demonstrated how a personalized multi-task approach to model mood, stress, and health of different participants is better than a single-task approach to model each participant separately \cite{taylor2017personalized}. 


\subsection{Ablation study}
\label{sec:results_ablation}

In this section we present our ablations studies. The goal of these experiments is to compare the performance of the visual and the text modalities, separately. For each modality we use the corresponding ST and MT architectures described in Sect.\ref{sec:emotion_modelling}, but we just use one input branch (text or visual). 

Tables \ref{tab:results_ablation_text} and \ref{tab:results_ablation_visual} show the results obtained for the separate text and visual modalities, respectively. For each of the two modalities we also provide, as a reference, the results obtained by the Baseline model for the corresponding modality.


\begin{figure*}[htb]
\centering
\begin{tabular}{@{}cccc@{}}
\hspace{-.23in}
\includegraphics[width=.52\textwidth]{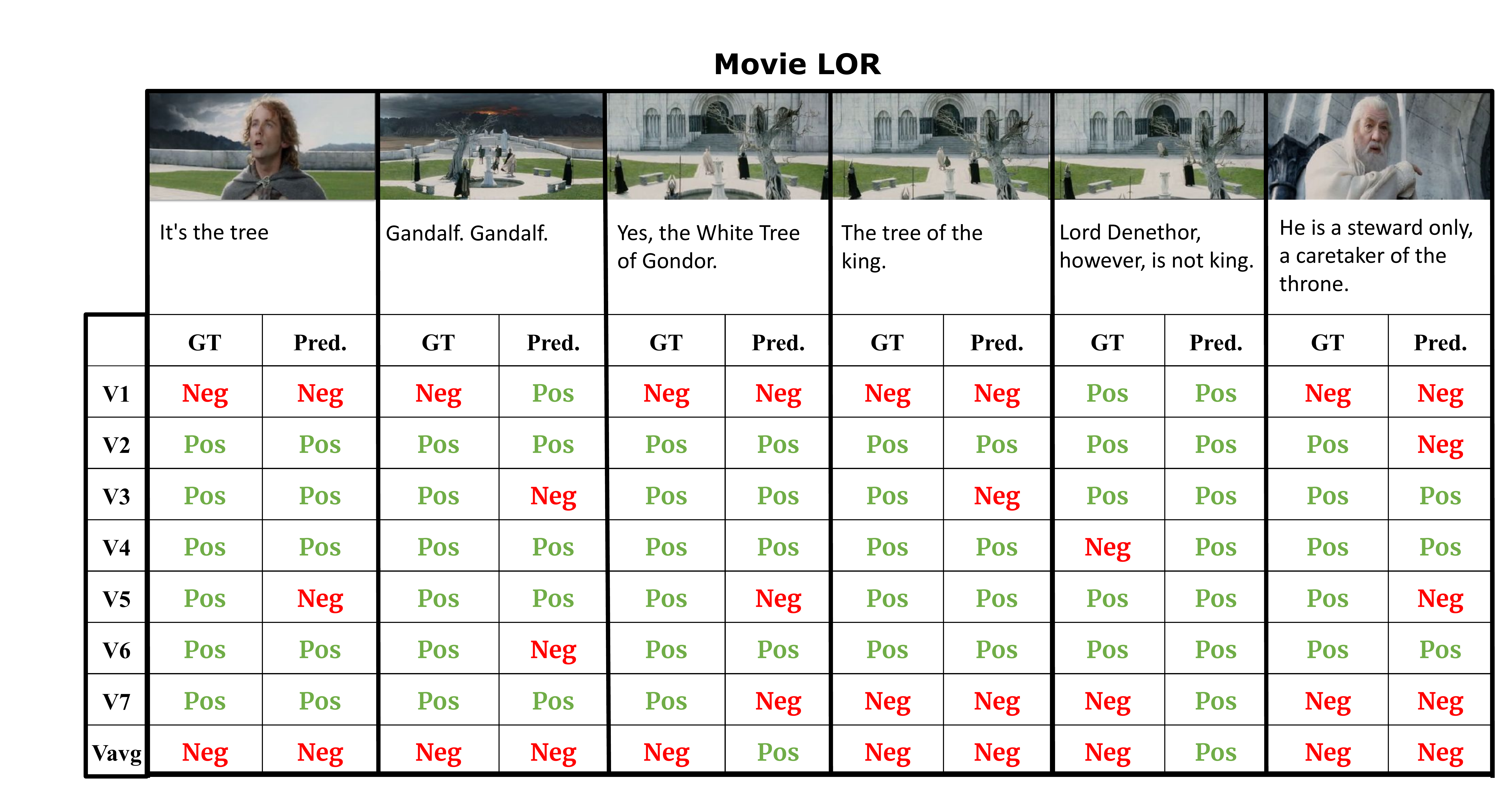} &
\hspace{-.26in}
\includegraphics[width=.52\textwidth]{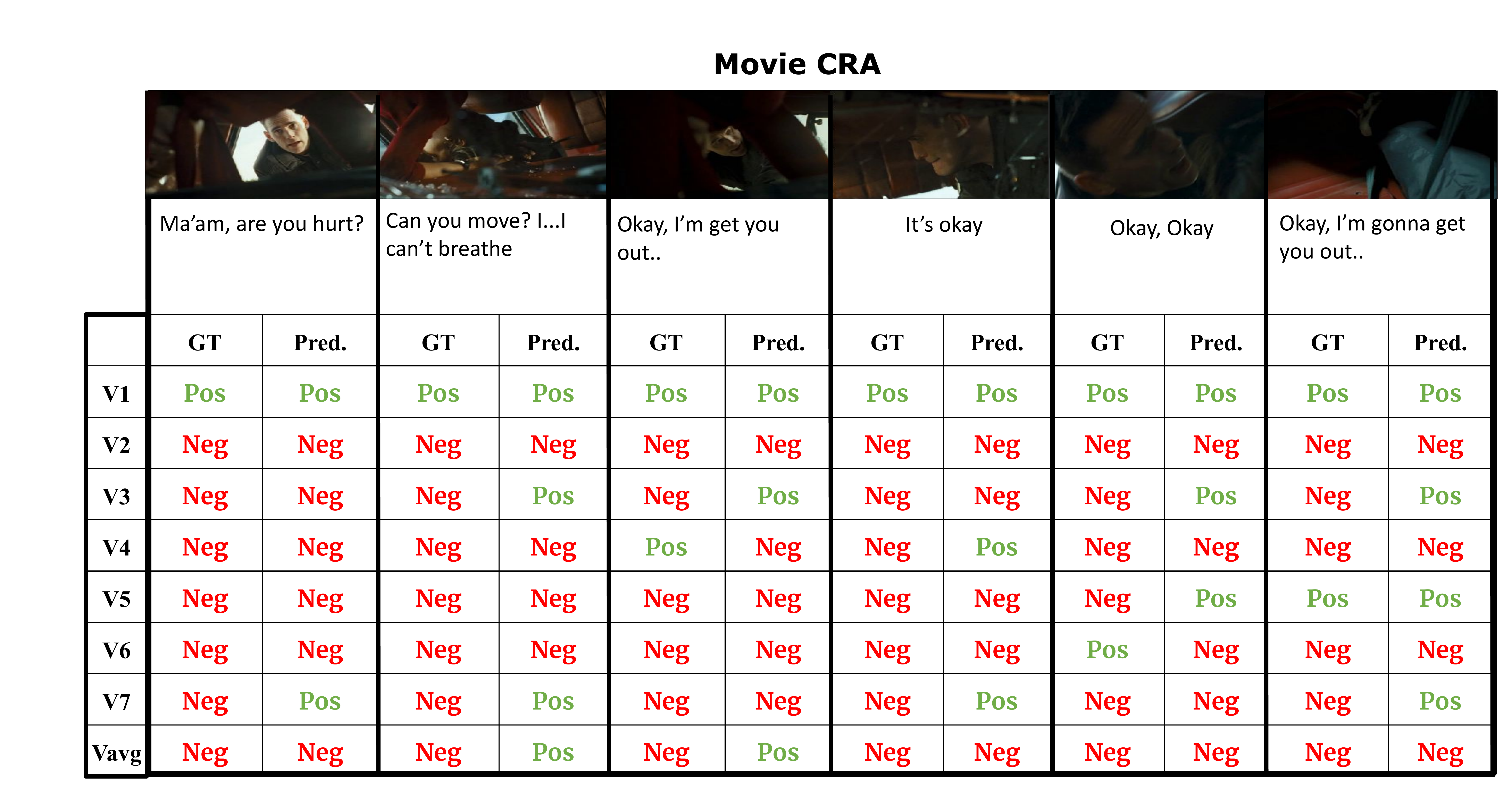} & \vspace{-.1in} \\
\hspace{-.23in}
\includegraphics[width=.52\textwidth]{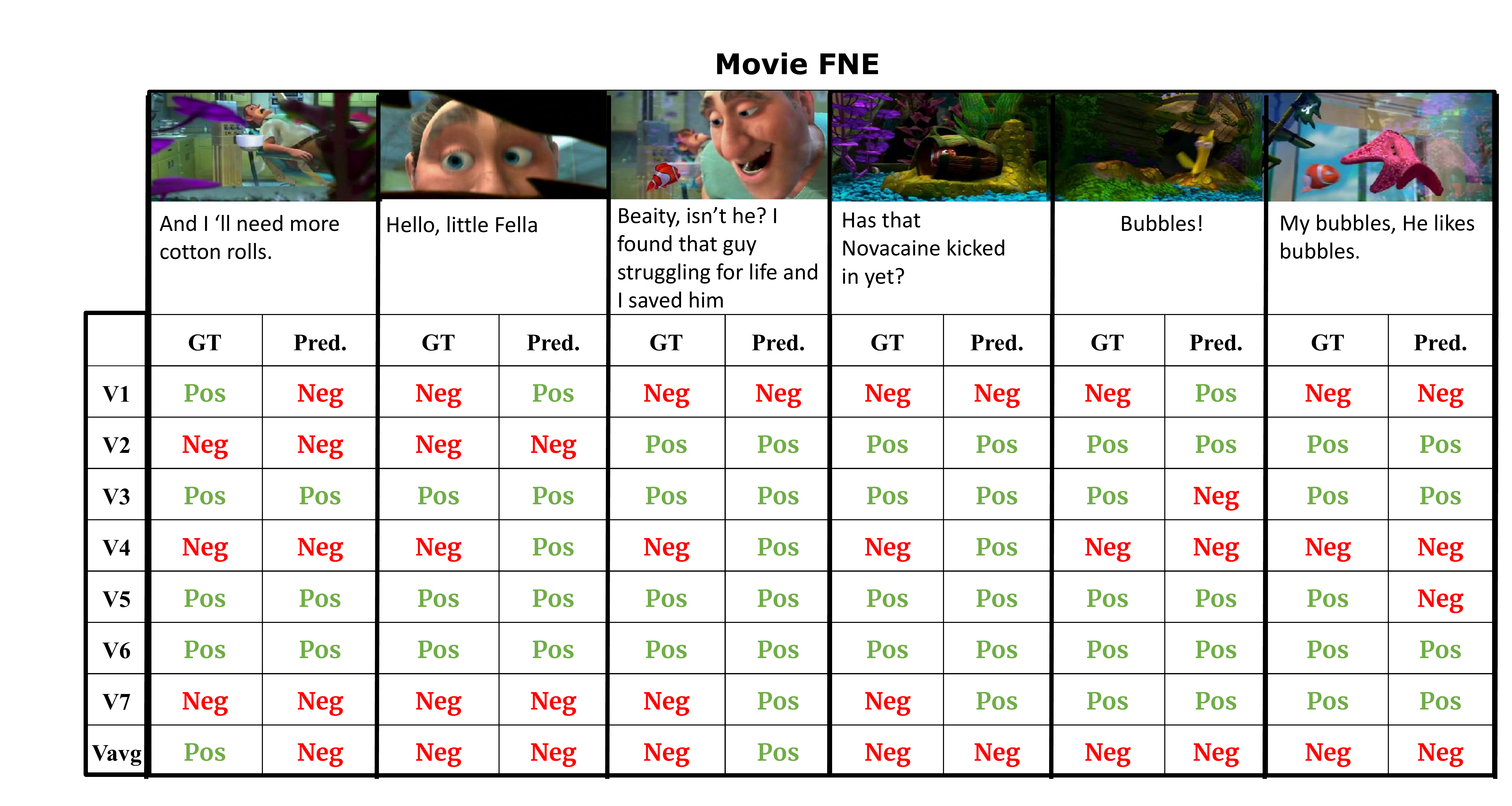} & 
\hspace{-.26in}
\includegraphics[width=.52\textwidth]{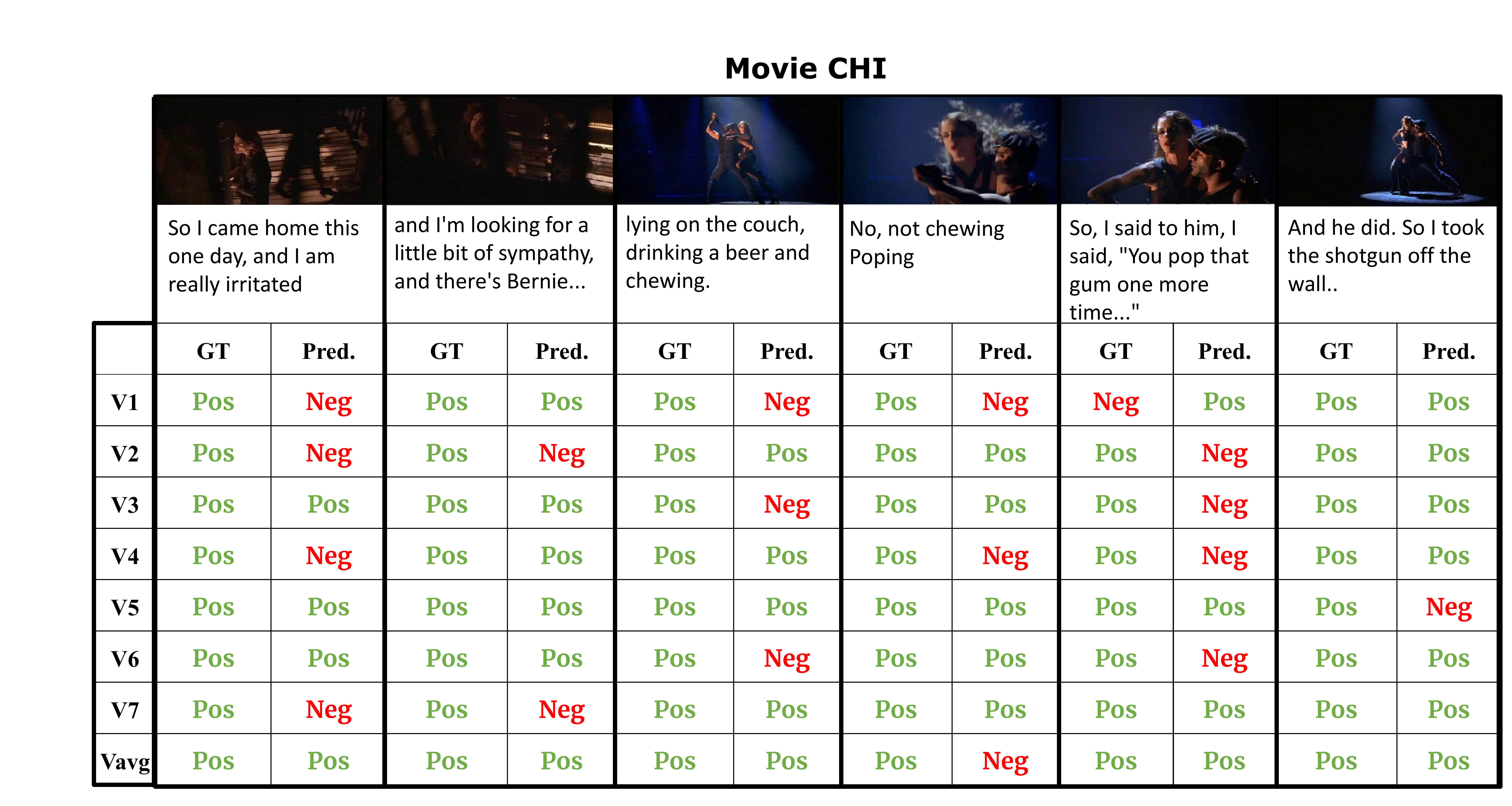} \\
\end{tabular}
\caption{Qualitative results. For some randomly selected segments of the movies LOR, CRA, FNE, and CHI, the figure shows the ground truth annotation per each viewer, including $V_{avg}$ (column GT) and the valence category predicted by our MT (multi-modal) model (column Pred.).}
\label{fig:results_qualitative}
\end{figure*}

For the text modality we observe that the Baseline approach is highly dependent on the data distribution. Concretely, we observed that most of the samples are classified as positive, which is the most represented class. Thus, when we experimented the Baseline-Text approach for every single viewer using cross-validation folds we got a significant drop in performance. For most of the viewers ($V_1$, $V_4$, $V_5$, $V_7$ and $V_{avg}$), the accuracy is less than the accuracy obtained by a random classifier (see Table \ref{tab:results_ablation_text}). We observe that our ST-Text and MT-Text architectures obtain better accuracies for the average viewer and for all the individual viewers except one ($V_3$). Notice that only the ST-Text model for $V_7$ gives an accuracy worse than the random.
Furthermore, Table \ref{tab:results_ablation_text} also shows that MT-Text architecture outperformed ST-Text in modeling all the viewers (52.96 vs. 65.04 mean accuracy) and also the average viewer (51.70 vs. 70.16). This supports our hypothesis that the MT approach suits better in modeling evoked emotions for multiple viewers rather than taking an average from multiple viewers first and then modeling the average evoked emotion.

In contrast with the text modality, the visual modality is richer and thus provides better results in terms of accuracy (see Table \ref{tab:results_ablation_visual}). We again tested Baseline-Visual for every single viewer and the viewer using cross-validation folds and found that our ST-Visual and MT-Visual approaches obtain better results than the Baseline-Visual. The results are shown in Table \ref{tab:results_ablation_visual}. We observe that the MT-Visual approach, when compared to the Baseline-Visual approach, is able to generalize better for each individual viewer plus the average viewer (66.60 vs. 70.10 mean accuracy) and average of all viewers (64.70 vs. 74.80).
\begin{table}[h] 
\caption{Comparison of the Baseline model with our proposed MT approach on the Baseline movie split. Accuracies obtained when modelling the average viewer ($V_{avg}$).}
    \begin{center}
    \begin{tabular}{|c|c|c|c|} 
      \hline
       & Text & Visual & Multimodal \\ 
       \hline\hline
        Baseline \cite{Nguyen2019} &  65.1 &  82.8  &  83.2 \\ 
        \hline
        MT (ours) & \textbf{75.48} & \textbf{84.88} & 
        \textbf{89.50} \\
      \hline
      \end{tabular}
      \label{tab:results_ablation_baselinesplit}
      \end{center}
\end{table}

\begin{table}[hbt!]
\caption{Per each viewer, accuracies obtained by our MT approach (multi-modal, text, and visual) on the Baseline movie split. Last row displays the average among the accuracies obtained on $V_1$, ..., $V_7$.}
    \begin{center}
    \begin{tabular}{|c|c|c|c|c|} 
      \hline
      & MT (ours) & MT-Text (ours) & MT-Visual (ours) \\
      \hline\hline
      $V_1$ & \textbf{74.30} & 63.45 & 72.86 \\
      \hline
      $V_2$ & \textbf{77.46} & 67.50 & 69.63 \\
      \hline
      $V_3$ & \textbf{76.37} & 68.70 & 70.27 \\
      \hline
      $V_4$ & \textbf{73.71} & 71.28 & 68.70\\
      \hline
      $V_5$ & \textbf{85.89} & 65.44 & 70.42 \\
      \hline
      $V_6$ & \textbf{71.52} & 70.82 & 76.90 \\
      \hline
      $V_7$ & \textbf{70.63} & 68.90 & 70.77 \\
      \hline
      Mean & \textbf{75.69} & 68.01 & 71.36 \\
      \hline
      \end{tabular}
      \label{tab:results_basline_split}
      \end{center}
\end{table}

\subsection{Comparing MT with the Baseline model}
\label{tab:results_baseline_viewer}
In this section we compare the performance of our best model (MT) with the performance of the Baseline multi-modal model \cite{Nguyen2019}. For this purpose we report accuracies on the data split used in \cite{Nguyen2019}, since the code for the Baseline multi-modal approach is not available and we could not train it in our fold of data partitions. Concretely, the data split of the Baseline model is the following: $5$ movies for training (BMI, CHI, FNE, GLA, LOR), and $2$ movies for testing (CRA, DEP). We compare Baseline accuracies with the accuracies we got from $V_{avg}$ branch of our MT-text, MT-visual, and MT-multimodal models. The results are shown in Tab.\ref{tab:results_ablation_baselinesplit}. We observe that our MT approach outperforms the Baseline in the text (65.1 vs 75.48) and the visual (82.8 vs 84.88) modalities. Our multi-modal approach, which uses just text and visual modalities, also outperforms the Baseline multimodal approach, which also uses the audio modality (83.2 vs 89.50).

Finally, Tab.\ref{tab:results_basline_split} shows for our MT models the accuracies obtained per viewer on the Baseline movie split. As before, we observe that the MT multi-modal model is the one obtaining the highest accuracies for all the viewers (75.69), when compared to MT models that use just the text modality (68.01) or just the visual modality (71.36). 

\subsection{Discussion}
\label{sec:results_discussion}

Our results show that the proposed MT approach to model the aggregated (average) viewer ($V_{avg}$) obtains the highest accuracies in the valence classification. This is observed in our cross-fold validation as well as in the data split reported in the Baseline model paper \cite{Nguyen2019}. Another important observation is that the MT approach obtains better accuracies in approximating each viewer with respect to the ST approach. Again, this is observed both in our cross-fold validation as well as in the data split of the Baseline model. In our ablation studies, we also observe the MT approach showing the highest accuracies for both the text and the visual modalities. Furthermore, we observe the visual modality achieves better results than the text modality, when the modalities are tested separately. A higher performance on the visual modality, when compared to the text modality, was also observed in \cite{Nguyen2019}. Interestingly, we also observe that in our results the worst performance, across all the viewers, is usually obtained by $V_1$. Notice that, as shown in Sect.\ref{Data distribution}, $V_1$ was actually the viewer that was less correlated with the others, as shown in Fig. \ref{Average correlation}. 

Finally, Fig. \ref{fig:results_qualitative} provides some qualitative results. Concretely, for four of the movies (LOR, CRA, FNE, CHI), we show the ground truth annotation per viewer (GT column) and the predictions made by our MT-multimodal model (Pred. column) for some segments. First, we observe how, for the same segment, different viewers provide different ground truth annoations. This qualitatively illustrates the observations we made in Sect.\ref{sec:viewer_correlations}, where we showed the histograms of the annotation distribution per viewer (Fig.\ref{Data distribution}). In these examples we also observe how $V_1$ often disagrees with the rest of the viewers, as shown also before in Sect.\ref{sec:viewer_correlations}, when we analyzed the viewers correlations (Fig.\ref{Average correlation}). Finally, we observe how the MT model can often approximate the different opinions of the viewers, while approximating the average viewer $V_{avg}$.

\section{conclusion}
In this paper we propose a Multi-Task (MT) Learning approach for modeling emotion evoked by movies. Our MT approach jointly models the evoked emotion per each viewer as well as the average evoked emotion. 
 Concretely, our architecture has two separate blocks: a shared convolutional block for all the viewers, with specific backbones for the visual and text modality, respectively, and a dedicated fully-convolutional block for each viewer, including the average viewer. We show that the accuracy obtained by this MT architecture is significantly higher than other techniques directly trained on the average viewer. 

For our experiments we use the COGNIMUSE dataset. Concretely, we perform an extended set of experiments by considering cross-validation instead of evaluating the models in a single data partition. Our results show that the proposed MT approach obtains the highest accuracies in the valence classification of the aggregated annotations. Furthermore, the MT approach also obtains better accuracies in approximating each single viewer. This is observed in our multimodal formulation as well as in our ablations studies that evaluate, separately, the visual and the text modalities.

We believe that the proposed MT approach can be useful for other tasks that need to deal with subjective judgements or perceptions, such as recommender systems or emotion recognition in other contexts. Also, we hope our results will encourage the release of individual annotations instead of just the aggregated annotations in future data collection efforts.

The code is available in our project repository\footnote{https://github.com/HassanHayat08/Recognizing-Emotions-evoked-by-Movies-usingMultitask-Learning}.

 \section*{Acknowledgments}
This work was partially supported by the Spanish Ministry of Science, Innovation and Universities and the European Regional Development Fund, RTI2018-095232-B-C22. HH is supported by a grant from the Universitat Oberta de Catalunya. We thank NVIDIA for their generous hardware donations.



\begin{thebibliography}{00}
\bibitem{russell2003core}
    J. A. Russell, “Emotion, core affect, and psychological construction,” Cognition \& Emotion, vol. 23, no. 7, pp. 1259–1283, Nov. 2009.
\bibitem{plantinga2013affective}
    C. Plantinga, “The Affective Power of Movies,” Psychocinematics, pp. 94–112, Mar. 2013.

\bibitem{plantinga2008emotion}
    C. Plantinga, “Emotion and Affect,” The Routledge Companion to Philosophy and Film, pp 106-116, 2008.

\bibitem{2016mediaeval}
    Dellandréa E, Chen L, Baveye Y, Sjöberg MV, Chamaret C, “The mediaeval 2016 emotional impact of movies task", InCEUR Workshop Proceedings, Nov 22, 2016.

\bibitem{sjoberg2015mediaeval}
    Sjöberg M, Baveye Y, Wang H, Quang VL, Ionescu B, Dellandréa E, Schedl M, Demarty CH, Chen L, “The MediaEval 2015 Affective Impact of Movies Task. InMediaEval", Sep 14, 2015.

\bibitem{soleymani2014corpus}
    Soleymani M, Larson M, Pun T, Hanjalic A. “Corpus development for affective video indexing", IEEE Transactions on Multimedia. vol. 16, pp. 1075-89, Feb 10, 2014.

\bibitem{thao2019multimodal}
	Thao HT, Herremans D, Roig G., “Multimodal Deep Models for Predicting Affective Responses Evoked by Movies," InICCV Workshops, pp. 1618-1627, Oct 1, 2019.

\bibitem{Poria_2017}
    Poria S, Peng H, Hussain A, Howard N, Cambria E., “Ensemble application of convolutional neural networks and multiple kernel learning for multimodal sentiment analysis," Neurocomputing, pp.217-230, Oct 25, 2017.

\bibitem{muszynski2019recognizing}
    M. Muszynski, L. Tian, C. Lai, J. D. Moore, T. Kostoulas, P. Lombardo, T. Pun, and G. Chanel, “Recognizing Induced Emotions of Movie Audiences from Multimodal Information,” IEEE Transactions on Affective Computing, vol. 12, no. 1, pp. 36–52, Jan. 2021.

\bibitem{Thao_2019}
    H. T. P. Thao, D. Herremans, and G. Roig, “Multimodal Deep Models for Predicting Affective Responses Evoked by Movies,” 2019 IEEE/CVF International Conference on Computer Vision Workshop (ICCVW), Oct. 2019.

\bibitem{timar2018feature}
    Timar Y, Karslioglu N, Kaya H, Salah AA., “Feature Selection and Multimodal Fusion for Estimating Emotions Evoked by Movie Clips," InProceedings of the 2018 ACM on International Conference on Multimedia Retrieval, pp. 405-412, June 5, 2018.

\bibitem{liu2019multimodal}
    Liu W, Qiu JL, Zheng WL, Lu BL., “Multimodal emotion recognition using deep canonical correlation analysis," arXiv preprint, arXiv:1908.05349, Aug 13,2019.
  
\bibitem{pini2017modeling}
    Pini S, Ahmed OB, Cornia M, Baraldi L, Cucchiara R, Huet B., “Modeling multimodal cues in a deep learning-based framework for emotion recognition in the wild," InProceedings of the 19th ACM International Conference on Multimodal Interaction, pp. 536-543, Nov 3, 2017.

\bibitem{zlatintsi2017cognimuse}
    Zlatintsi A, Koutras P, Evangelopoulos G, Malandrakis N, Efthymiou N, Pastra K, Potamianos A, Maragos P., “COGNIMUSE: A multimodal video database annotated with saliency, events, semantics and emotion with application to summarization. EURASIP Journal on Image and Video Processing," vol. 1, pp.1-24, Dec 2017.
    
\bibitem{zheng2017identifying}
    W.-L. Zheng, J.-Y. Zhu, and B.-L. Lu, “Identifying Stable Patterns over Time for Emotion Recognition from EEG,” IEEE Transactions on Affective Computing, vol. 10, no. 3, pp. 417–429, Jul. 2019.
 
\bibitem{becker2017emotion}
    H. Becker, J. Fleureau, P. Guillotel, F. Wendling, I. Merlet, and L. Albera, “Emotion Recognition Based on High-Resolution EEG Recordings and Reconstructed Brain Sources,” IEEE Transactions on Affective Computing, vol. 11, no. 2, pp. 244–257, Apr. 2020.

\bibitem{wang2019sentidiff}
    L. Wang, J. Niu, and S. Yu, “SentiDiff: Combining Textual Information and Sentiment Diffusion Patterns for Twitter Sentiment Analysis,” IEEE Transactions on Knowledge and Data Engineering, vol. 32, no. 10, pp. 2026–2039, Oct. 2020.
  
\bibitem{rosas2013multimodal}
    V. Perez Rosas, R. Mihalcea, and L.-P. Morency, “Multimodal Sentiment Analysis of Spanish Online Videos,” IEEE Intelligent Systems, vol. 28, no. 3, pp. 38–45, May 2013.
  
\bibitem{chang2017learning}
    J. Chang and S. Scherer, “Learning representations of emotional speech with deep convolutional generative adversarial networks,” 2017 IEEE International Conference on Acoustics, Speech and Signal Processing (ICASSP), Mar. 2017.

\bibitem{Nguyen2019}
    Nguyen TL, Kavuri S, Lee M., “A multimodal convolutional neuro-fuzzy network for emotion understanding of movie clips," Neural Networks, vol 118, pp208-219, Oct 1, 2019.

\bibitem{ghaleb2019multimodal}
    E. Ghaleb, M. Popa, and S. Asteriadis, “Multimodal and Temporal Perception of Audio-visual Cues for Emotion Recognition,” 2019 8th International Conference on Affective Computing and Intelligent Interaction (ACII), Sep. 2019.
    
\bibitem{ko2018brief}
    B. Ko, “A Brief Review of Facial Emotion Recognition Based on Visual Information,” Sensors, vol. 18, no. 2, p. 401, Jan. 2018.
    
\bibitem{jan2016bul}
    Jan A, Gaus YF, Meng H, Zhang F. BUL, “MediaEval 2016 Emotional Impact of Movies Task," InMediaEval, Oct 20, 2016.
  
\bibitem{ma2016thu}
    Ma Y, Ye Z, Xu M., “THU-HCSI at MediaEval 2016: Emotional Impact of Movies Task," InMediaEval, Oct 20, 2016.


\bibitem{bao2013your}
    Bao X, Fan S, Varshavsky A, Li K, Roy Choudhury R., “Your reactions suggest you liked the movie: Automatic content rating via reaction sensing," InProceedings of the 2013 ACM international joint conference on Pervasive and ubiquitous computing, pp. 197-206, Sep 8, 2013.

\bibitem{lee2014emotion}
    G. Lee, M. Kwon, S. Kavuri Sri, and M. Lee, “Emotion recognition based on 3D fuzzy visual and EEG features in movie clips,” Neurocomputing, vol. 144, pp. 560–568, Nov. 2014.

\bibitem{fayek2016modeling}
    Fayek HM, Lech M, Cavedon L., “Modeling subjectiveness in emotion recognition with deep neural networks: Ensembles vs soft labels," In2016 international joint conference on neural networks (IJCNN), pp.566-570,Jul 24, 2016.

\bibitem{chou2019every}
    Chou HC, Lee CC., “Every rating matters: Joint learning of subjective labels and individual annotators for speech emotion classification," IEEE International Conference on Acoustics, Speech and Signal Processing (ICASSP),
    pp. 5886-5890, May 12, 2019.

\bibitem{li2019attentive}
    Li JL, Lee CC., “Attentive to Individual: A Multimodal Emotion Recognition Network with Personalized Attention Profile," InInterspeech, pp. 211-215, Sep, 2019.

\bibitem{han2017hard}
    J. Han, Z. Zhang, M. Schmitt, M. Pantic, and B. Schuller, “From Hard to Soft,” Proceedings of the 25th ACM international conference on Multimedia, Oct. 2017.

\bibitem{douglas2007humaine}
    E. Douglas-Cowie, R. Cowie, I. Sneddon, C. Cox, O. Lowry, M. McRorie, J.-C. Martin, L. Devillers, S. Abrilian, A. Batliner, N. Amir, and K. Karpouzis, “The HUMAINE Database: Addressing the Collection and Annotation of Naturalistic and Induced Emotional Data,” Lecture Notes in Computer Science, pp. 488–500.

\bibitem{schaefer2010assessing}
    A. Schaefer, F. Nils, X. Sanchez, and P. Philippot, “Assessing the effectiveness of a large database of emotion-eliciting films: A new tool for emotion researchers,” Cognition \& Emotion, vol. 24, no. 7, pp. 1153–1172, Nov. 2010.
  
\bibitem{baveye2015liris}
    Y. Baveye, E. Dellandrea, C. Chamaret, and Liming Chen, “LIRIS-ACCEDE: A Video Database for Affective Content Analysis,” IEEE Transactions on Affective Computing, vol. 6, no. 1, pp. 43–55, Jan. 2015.


\bibitem{ye_subjective_annotations}
    J. Ye, J. Li, M. G. Newman, R. B. Adams, and J. Z. Wang, “Probabilistic Multigraph Modeling for Improving the Quality of Crowdsourced Affective Data,” IEEE Transactions on Affective Computing, vol. 10, no. 1, pp. 115–128, Jan. 2019.

\bibitem{kingma2014adam}
    Kingma DP, Ba J., “Adam: A method for stochastic optimization”, arXiv preprint arXiv:1412.6980,
    Dec 22, 2014.

\bibitem{Tibshirani1996}
    R. Tibshirani, “Regression shrinkage and selection via the lasso: a retrospective,” Journal of the Royal Statistical Society: Series B (Statistical Methodology), vol. 73, no. 3, pp. 273–282, Apr. 2011.

\bibitem{Carreira_2017}
    J. Carreira and A. Zisserman, “Quo Vadis, Action Recognition? A New Model and the Kinetics Dataset,” 2017 IEEE Conference on Computer Vision and Pattern Recognition (CVPR), Jul. 2017.
    
\bibitem{ioffe2015batch}
    Ioffe, Sergey, and Christian Szegedy., “Batch normalization: Accelerating deep network training by reducing internal covariate shift,” In International conference on machine learning, pp. 448-456. 2015.

\bibitem{taylor2017personalized}
    Taylor SA, Jaques N, Nosakhare E, Sano A, Picard R., “Personalized Multitask Learning for Predicting Tomorrow’s Mood. Stress, and Health,” IEEE Transactions on Affective Computing, vol.8, no. 8,
    pp. 17-33, Jun. 2017. 

\end{thebibliography}
\end{document}